# Exploring AI-Generated Text in Student Writing: How Does AI Help?

David James Woo [a, *], Hengky Susanto [b], Chi Ho Yeung [b], Kai Guo [c], and April Ka Yeng Fung [d]

[a] Precious Blood Secondary School, Hong Kong
[b] Department of Science and Environmental Studies, The Education University of Hong Kong, Hong Kong
[c] Faculty of Education, The University of Hong Kong, Hong Kong
[d] Hoi Ping Chamber of Commerce Secondary School, Hong Kong

[*] Corresponding author
  - Email address: net_david@pbss.hk
  - Postal address: Precious Blood Secondary School, 338 San Ha Street, Chai Wan, Hong Kong

**Declarations of interest**
None.

**Acknowledgement**
The work by C.H.Y. is supported by the Research Grants Council of the Hong Kong Special Administrative Region, China (Projects No. EdUHK GRF 18301119).

# Exploring AI-Generated Text in Student Writing: How Does AI Help?

**Abstract**

English as a foreign language (EFL) students' use of artificial intelligence (AI) tools that generate human-like text may enhance students' written work. However, the extent to which students use AI-generated text to complete a written composition and how AI-generated text influences the overall writing quality remain uncertain. 23 Hong Kong secondary school students wrote stories with AI-writing tools, integrating their own words and AI-generated text into the stories. We analyzed the basic structure, organization, and syntactic complexity of each story and its AI-generated text. Experts scored the quality of each story's content, language, and organization. By employing multiple linear regression and cluster analyses, we found that both the number of human words and the number of AI-generated words significantly contributed to writing scores. Furthermore, students could be classified into competent and less competent writers based on the variations of students' usage of AI-generated text compared to their peers. Cluster analyses revealed some benefit of AI-generated text in improving the scores of both high-scoring students' and low-scoring students' writing. We suggest differentiated, pedagogical strategies for EFL students to effectively use AI-writing tools and AI-generated text to complete writing tasks.

*Language(s) Learned in This Study:* English

**Exploring AI-Generated Text in Student Writing: How Does AI Help?**

English as a foreign language (EFL) students can face significant challenges in the writing classroom. Compared to L1 students, they typically have an inferior linguistic knowledge base and their written texts tend to be less cohesive and fluent, more error prone and shorter (Hyland, 2003a). They often do not have access to cultural resources that L1 students have and that enable discovery of genre outcomes (Hyland, 2003b). Besides, there is evidence showing students do not receive sufficient writing instruction and feedback from teachers (Butterfuss et al., 2022).

The implementation of language models or machine-learning systems trained on millions of webpages (Radford et al., 2019) may be one way to address EFL students' challenges in the writing classroom. This is not only because language models accurately perform a variety of language tasks but also because they are capable of artificial intelligence (AI) natural language generation (NLG), that is, generating coherent, lengthy text, indistinguishable from human writing (Brown et al., 2020). Studies have shown students' interaction with AI-NLG tools can benefit their writing. For example, Dizon and Gayed (2021) found that Japanese university students produced more lexical variation and fewer grammatical errors when they wrote with Grammarly, a predictive text and intelligent writing commercial application that is integrated into word processing applications. Kangasharju et al. (2022) designed a Poem Machine that drafted poems that students could not only take as inspiration, but also edit and revise so as to compose poems written with AI-generated text. Importantly, their data suggested an association between the quality of a student's poem and the number of edits made on the draft poem. Thus, just as students increasingly consult online sources and repurpose information from those sources for their writing (Tan, 2023), so students may benefit from consulting AI-NLG tools as a language resource with the specific aim of integrating text from the tools into students' written compositions.

To advance the use of AI-NLG tools as a collaborative tool and language resource in the EFL writing classroom, the present study is interested in exploring students' usage of AI-generated text in written compositions. We are particularly interested in exploring how AI-generated text may affect the overall quality of compositions. By understanding this phenomenon, we may inform EFL writing instruction for using AI-NLG tools and their generative texts.

**Human-AI Collaborative Writing Process**

We view writing from a sociocultural perspective (Vygotsky, 1978). Writing is an activity situated within a context and mediated by tools (Prior, 2006). Within a typical EFL classroom with teachers and peers, texts have mediated students' writing activity and subsequently, digital tools. Not all tools may effectively mediate a learner's writing development because the tools do not meet that learner's needs (Vygotsky, 1978). Besides, learners may have lacked strategies to navigate the tools and other elements of their socio-technical environment to improve the quality of their writing (Crossley et al., 2016).

AI-NLG is an emergent type of digital tool and we frame its mediation on EFL students' writing development as writing with a *machine-in-the-loop*, which emphasizes the collaboration between a human writer and an AI-NLG tool (Clark et al., 2018) as each contributes text so as to complete a written composition.

Machine-in-the-loop studies have contributed knowledge on different samples of human writers, the degree of agency the human writer has when collaborating with the AI-NLG tool, and different measures of writing quality. Clark et al. (2018) found that the creativity, coherence and grammatical accuracy of stories by adults who wrote with un-editable suggestions of AI-generated text was no better than those by adults who wrote without those suggestions. Calderwood et al. (2020) found four professional novelists preferred writing stories with an AI-NLG tool that generated short, editable chunks of text to writing with a tool that generated long, un-editable chunks of text in the novelists' stories. Yang et al. (2022) piloted a turn-taking approach to writing with an AI-NLG tool. University students and the tool would take turns writing paragraphs of a story, with the students able to edit AI-generated paragraphs or to regenerate AI-paragraphs. The researchers found the students preferred the former ability of the tool. In analyzing the written compositions, they found that AI-generated text was largely coherent with human text and that AI-generated text could contribute creative twists to stories. Gayed et al. (2022) developed an AI-NLG tool that, when prompted, would generate possible next words for the prompt with confidence scores for the next words. They found their adult EFL students could output a greater number of words with greater syntactic complexity when writing with the tool to complete a timed-writing task than when writing without the tool. However, they did not find a statistically significant relationship between writing with the tool and higher lexical diversity in students' written compositions.

      From these machine-in-the-loop studies, it appears that adults prefer writing with AI-NLG tools that are non-intrusive and generate editable text. It also appears that compositions written with AI-generated text show mixed results in different measures of writing quality. However, previous studies appear to have measured the quality of written compositions without having captured and analyzed the AI-generated text in the compositions. Thus, we do not know how people strategically edited text from AI-NLG tools, interpolating it into written work, and how different patterns of AI-generated text in compositions may contribute to writing quality.

The strategic use of AI-generated text in compositions may lead to higher quality writing, but there is a lack of empirical evidence on interpolation of AI-generated text in written compositions, for example, in terms of the number of AI-generated words in a composition, the number of instances of AI-generated text and other statistics about these instances. With that evidence, patterns of how writers use AI-generated text may be identified and associated with the perceived quality of written compositions.

**This Study**

We build on existing machine-in-the-loop studies to conceptualize our study. In our conceptual framework, the human writer and AI-NLG tool do not equally contribute text to a written work. Instead, the human writer exercises full agency and the AI-NLG tool plays a supporting role to complete a written composition, as the tool only generates text at a human writer's prompting. Besides, the human writer can ignore the AI-generated text. Otherwise, the writer can use and edit any of the text for a composition. Figure 1 illustrates our study's conceptualization as, first, a student provides an input prompt to an AI-NLG tool. After the tool generates text that is a prediction of text that follows the prompt, the student evaluates this output and decides whether to integrate any of it into their composition. This cycle repeats until the student completes the composition. Like other studies, ours focuses on story writing and analyzing the quality of writing based on a completed composition.

**Figure 1**

*Machine-in-the-loop*

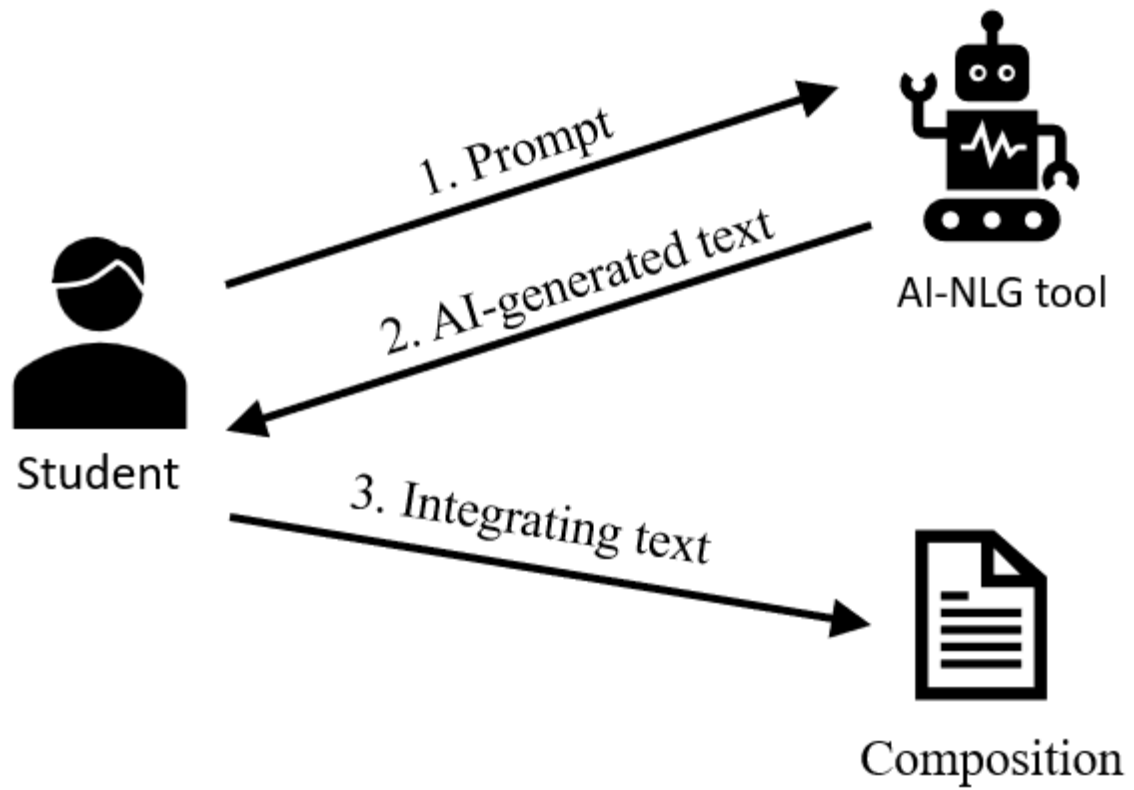

Our study explores writing with a machine-in-the-loop in the context of EFL secondary school students. We aim to collect data on the language features of AI-generated text in written compositions as language features in a text can reflect an EFL student's knowledge and facilitate their growth (Crossley, 2020). With data on AI-generated text in compositions, we aim to identify any patterns of how students have strategically edited AI-generated text into their compositions. As we are interested in whether any patterns affect the quality of students' written compositions, we score their compositions written with AI-generated text and employ a multiple linear regression analysis. We employ a cluster analysis to identify any salient learner profiles, which are distinct student groups, each of which has achieved a particular quality of composition and has utilized AI-generated text in a particular way to complete compositions. Through the latter analysis we will have a better understanding as to how the use of AI-generated, English language text may improve the composition quality of different types of learners, if at all. By identifying any benefit from using AI-generated text on students' compositions, we hope to inform teaching and learning practice in the EFL classroom. The study was guided by the following three research questions (RQs):

**RQ1**: What are the language features of AI-generated text in students' compositions written with AI?

**RQ2**: What patterns of interaction with AI-generated text can be identified in students' compositions and what are their differences?

**RQ3**: How do interaction patterns with AI-generated text impact and benefit different types of learners' writing, if at all?

## Methodology

**Research Participants and Context**

This study was conducted purposefully in two Hong Kong secondary schools so as to sample students at different levels of English as a foreign language (EFL) achievement. One school, Ho Man Tin (HMT) School (pseudonym) receives primary school students at the 88-99th percentile of academic achievement in its geographic district. The other school, Chai Wan (CW) School (pseudonym) receives primary school students at the 44-55th percentile of academic achievement in its geographic district. Students attended two workshops at their respective schools presented by the first author from December 2022 to January 2023.

In the first workshop, students learned to code their AI-NLG tool. They used free-to-use resources: Python programming language, Gradio software development kit, and Hugging Face, a repository for machine-learning (ML) models and applications. A language model forms the basis of their AI-NLG tool and, initially, students were taught to design an AI-NLG tool by using one language model, GPT2, with one textbox for human text input and one text box of AI-generated text output. Upon successfully coding their first AI-NLG tool, students were then taught to swap GPT2 with other language models available on Hugging Face. Although the Hugging Face language models are smaller than the largest, proprietary language models such as ChatGPT, which can exceed 100 billion parameters (Brown et al., 2020), they varied widely in size, from hundreds of millions of parameters to GPT-J's six billion parameters (Wang & Komatsuzaki, 2021). For instance, Figure 2 shows a prompt and the output of an AI-NLG tool composed from the GPT-NEO 1.3 billion parameter language model. Finally, students were taught to design an AI-NLG tool by using more than one language model so that the tool could comprise several text boxes of AI-generated text output (see Figure 3). In having a choice of different models and number of outputs, students could access a greater number of outputs to integrate into their writing. Besides, students could compare model outputs and learn that the size of a language model can influence its performance in a variety of language tasks (Radford et al., 2019), and also the amount of computing resources and time that it needs to generate text (Simon, 2021). For instance, in Figure 2 we can see GPT-NEO-1.3B produced text lengths of up to several sentences unlike the language models in Figure 3. And in Figure 3, we can see the language models have different understandings of how to complete the same prompt. Nonetheless, the output of Hugging Face language models used in Figures 2 and 3 showed proper English language capitalization, punctuation, spacing and paragraphing.

**Figure 2**

*An AI-NLG tool comprising one language model and one textbox of output*

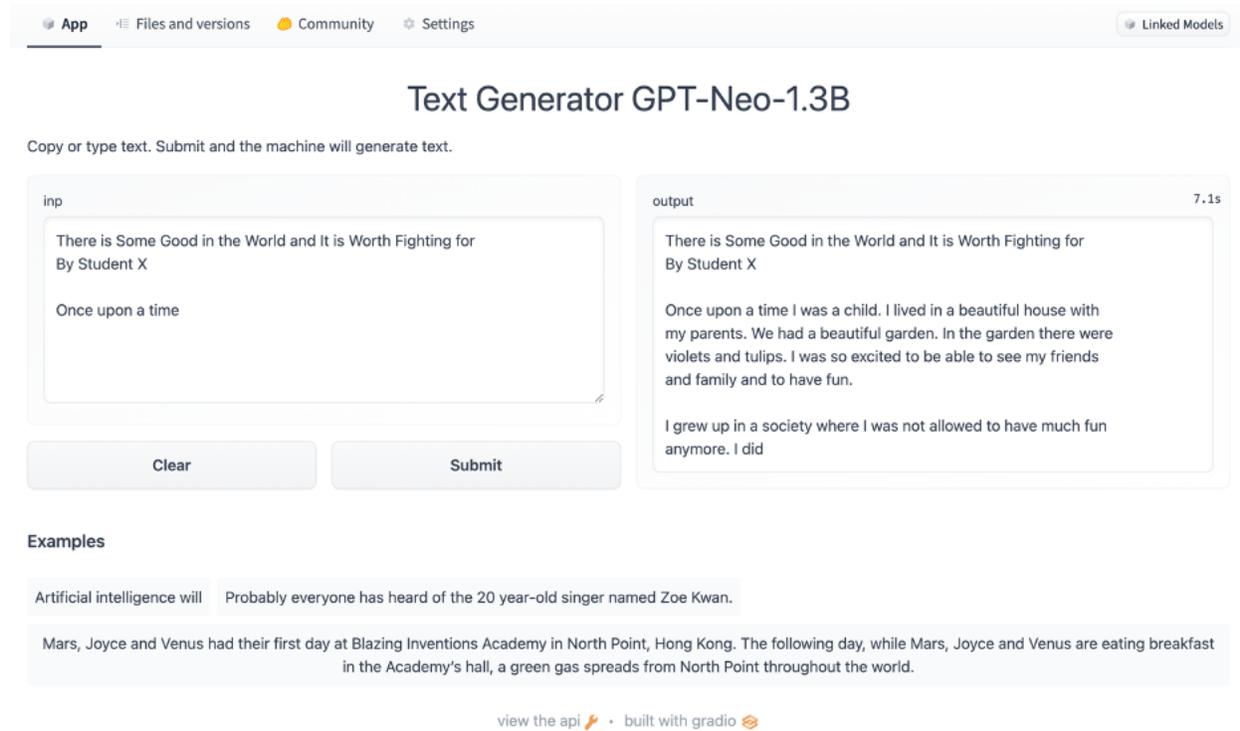

**Figure 3**

*An AI-NLG tool comprising three language models and three textboxes of output*

In the second workshop, students learned to interact with their AI-NLG tool by writing with a machine-in-the-loop (Clark et al., 2018). Students wrote with the support of their AI-NLG tool and with full autonomy to act on AI's output, if at all, so that their creativity might be enhanced (Singh et al., 2022). To support this approach, students were taught zero-shot prompting methods (Brown et al., 2020) such as prompting a tool with different lengths of story text, natural language instruction and questions. They were also taught digital writing skills for selecting AI-generated text, and editing that text into a story. Students used their AI-NLG tools for 45 minutes. They could use the tools freely and repeatedly at that time. Finally, students were introduced to a contest where they could submit their short stories to the contest organizer, and stories would be scored according to a rubric to determine winners. Students did not need to complete their short stories for contest submission during that time. Students were informed that their completed stories would be collected and analyzed for scientific purposes, their stories would be anonymized for study and they had the right to decline participation at any stage of study.

**Data Collection**

Students wrote stories on Google Docs and shared their documents with the first author. Any story that exceeded the contest's 500-word limit was not included in this study. Any story that appeared incomplete but fell within the 500-word limit was included in this study. Thus, this study collected 23 stories, 16 from HMT School students and seven from CW School students.

Since each story comprises a student's own words and AI-generated text, to facilitate our analysis of a student's interaction patterns with AI-generated text, a student highlighted the student's own text in red and AI-generated text in black (see Figure 4).

**Figure 4**

*A story written on Google Docs with a student's words and AI-generated text*

<div style="color:center">Mistake?</div>

<span style="color:red">On a chilly winter night, Maddison enters her bedroom at midnight and falls upon what she thinks is a pen.</span> She picks it up and it turns out to be… a crystal? She said , "woah! This crystal is ginormous ! where did this crystal come from?"

<span style="color:red">Soon, the crystal</span> starts glowing a bright purple, <span style="color:red">a bright light flashes her eyes and teleports her to an unknown realm.</span> Then the light turns into an orb of energy. The orb introduces itself as the crystal king, he wants to hire maddie as <span style="color:red">his right hand to take over Lego mania.</span> The orb gives maddison a choice: join him. After a moment, she agrees. Then, The crystal king also gives her a gift, <span style="color:red">a crystal.</span>

<span style="color:red">After Maddison awakens, the crystal tells her to go to her</span> old boarding school and plant the crystal in the <span style="color:red">bonsai plant</span> on the front counter. When she woke up, she then witnessed the <span style="color:red">crystal</span> grow bigger and bigger eventually turning the school into a fortress.

<span style="color:red">Soon, six colour coded ninja come, the green one steps forward and asks. 'Harumi how are you alive!? I saw you get crushed in a crumbling building. He</span> said that there was so much rubble that he couldn't find you after days and nights. <span style="color:red">Maddison then answers with "what are you talking about? I'm Maddison, who is harumi?'</span> Everyone had a shocked look on their face <span style="color:red">by that statement.</span> You continue saying <span style="color:red">'found this crystal in my room and this orb named the crystal king came, hired me to help him take over somewhere called Lego mania'.</span> Wait, so you aren't the hired villain for <span style="color:red">Harumi?</span> Asked <span style="color:red">the blue ninja. Maddison</span> wonders if she looks that much different with eye makeup. 'Heavens no! I'm an ordinary school <span style="color:red">girl</span> from <span style="color:red">Chicago'</span> replied <span style="color:red">Maddison.</span>

<span style="color:red">With that, the crystal king snapped back with</span> 'I don't hire children as villians, I'm going to send you back' <span style="color:red">A spark of light appeared from the crystal king's hand and Maddison was transported from Lego Mania to her bedroom.</span>

<span style="color:red">Maddison</span> jumps into her bed with a feeling of deja vu, not knowing what just happened. She eventually fell asleep.

Note.
Text in red color indicates a student's own words and text in black indicates AI-generated text

**Scoring of Stories**

To prepare each story for human scoring, we removed a student's name and indicators of a student's text and AI-generated text. EFL subject matter experts scored: CW School's Native English Teacher (NET); CMT School's English panel head; and a postgraduate student researching EFL writing. To establish reliability in human scoring, the NET prepared a scoring rubric (see Appendix A) adapted from a standardized rubric that is used to assess the quality of English language writing in Hong Kong secondary schools and is familiar to the human scorers. The scorers were instructed to score stories for content (C) (e.g., creativity, and task completion), language (L) (e.g., grammar, punctuation, and spelling) and organization (O) (e.g., idea development and cohesiveness). The full mark for each criterion was five and a score was awarded in increments of one. Besides providing annotated examples of stories, the NET briefed the other two scorers and wrote notes for the awarding of specific marks (see Appendix A). These include awarding a score of one for C if a story was incomplete, and a story's L and O scores not exceeding the C score plus/minus one. Two experts independently marked the stories and we calculated the proportion of agreement (DeCuir-Gunby et al., 2010). The proportion of agreement for C was 73%, L 87% and O 70%. For all but two of the 23 stories, the scorers produced either the exact same CLO score, that is, the sum of C, L and O scores, or showed a CLO score difference of only one mark (see Appendix B). To reconcile any possible differences in scoring, we averaged the scores awarded by the two experts.

    The scoring rubric included an AI Words criterion with the scoring descriptors (see Appendix A). Like with C, L and O criteria, the AI Words scoring descriptors are categorized into different levels of performance, with the use of AI-generated text being more limited at the lower performance level (i.e. AI words in at least one chunk and at least one short chunk), more extensive at a higher level (i.e. at least eight chunks of AI words) and most extensive at the highest level (i.e. no more than 33% AI words). Additionally, a story needs to exhibit the descriptor at the lower performance level before it can be evaluated at a higher performance level. These descriptors were developed from our literature review and a pilot study of four students' stories written using their own words and AI-generated text. They were developed from measures of basic structure and organization and syntactic complexity that were observed in the stories with higher CLO scores. The NET scored de-anonymized versions of students' stories for the AI Words criterion.

By adding the CLO and AI words scores, we arrived at a grand total score for each story.

**Data Analysis**

To analyze language features of student's stories written with NLG tools, we operationalized the measures for the basic structure and organization of a composition and the syntactic complexity of AI-generated text as set out in the scoring rubric. Therefore, we sought to count the number of words in a story, and the number of AI-generated words and the number of student words. Furthermore, we sought to count the number of AI-generated text instances or chunks in a story. An AI chunk is defined as AI-generated text embedded within students' own text instances or chunks.

Syntactic complexity in EFL writing has referred to the variation and sophistication of production units or grammatical structures and is often measured in terms of length of production units (e.g. clauses, sentences) (Lu, 2010). Thus, to analyze the syntactic complexity of AI-generated text, we operationalized AI chunks into three syntactic forms based on sentence-length, a commonly measured syntactic unit in EFL writing (Hyland, 2003a), In our analysis, we found instances of AI-generated text that were shorter than five words in length but no instances of an AI-generated sentence that was shorter than five words. Therefore, we defined 1) a short AI chunk to be AI-generated text shorter than five words in length, and 2) a medium AI chunk to be AI-generated text longer than or equal to five words in length or sentence length. For example, an instance of an AI-generated punctuation mark would be categorized as a short AI chunk. Since we found instances of AI-generated text longer than sentence length, we defined those as long AI chunks. The NET manually counted the number of AI chunks and categorized each chunk for each story.

We prepared on Excel the descriptive statistics for the basic organization and structure and syntactic complexity measures. In *phase 1* of our findings, we examine these descriptive statistics alongside human-rated scores.

For insights into what patterns of interactions with AI-generated text might be more effective for human ratings of writing, we used statistical methods to compare scores from human raters with basic organization and structure and syntactic complexity statistics.

In *phase 2*, we analyzed the data using Multiple Linear Regression (MLR; Aiken et al., 2003), a statistical technique that describes the relationship between multiple variables, while taking the effect of each variable into account. This analysis reveals the factors that contribute to a student submission with a high score. For example, the algorithm explored the statistical relationship between students' score, amount of AI-generated words used in their writing, and the actual length of their writing, while exposing how these variables might affect the quality of the students' writing outcome.

Lastly, *in phase 3* we focused on discovering what salient learner profiles might emerge from the data and analyzing the effects of how AI's assistance might improve each type of learners' writing quality. Here, we performed cluster analysis using unsupervised machine-learning techniques Expectation-Maximization (EM) Algorithm (Dempster et al., 1977), K-Means Clustering (Macqueen, 1967), and Mean-Shift Clustering (Fukunaga et al., 1975).

## Results and Discussion

In this section, we first present and discuss the descriptive statistics of language features and scores. Second, we present and discuss the results of the multiple linear regression analysis and finally, the cluster analysis.

**Phase 1: Descriptive Statistics of Language Features and Scores**

As shown in Table 1, 21 out of the 23 stories contain AI chunks, and 19 of them contain long AI chunks but only 13 of them contain short AI chunks. If one considers that embedding shorter AI chunks into the stories is more granular editing between the AI-generated and human-written text, this result may imply that granular editing is challenging as fewer students achieved this goal. This is also because given the same length of stories, students would need to independently generate more human-written text to integrate shorter AI chunks. In addition, each story comprised an average of 3.77, 3.16 and 2.21 short, medium and long AI chunks, respectively, and each such story contained an average of 6.67 AI chunks. All these measures show a large standard deviation, with a coefficient of variation (i.e. the standard deviation divided by the average) equal to or higher than 70%, implying that there is a large variation in students' intention in embedding AI-generated words in their stories.

Moreover, Table 1 shows the average number of AI-generated words, human written words and the total number of words in the stories are 81.57, 248.57 and 323.04 respectively, and the standard deviation in all these measures is large. The large variation in the number of AI-generated words may imply that students vary a lot in their intention to incorporate AI-generated text. To better understand this variation, we show the distribution of the percentage of AI-generated words in Figure 5. From Table 1, although the average percentage of the usage of AI words is roughly 28%, and the majority of students used less than 20% of AI words in their stories, some students embedded a large percentage of AI words, up to almost 90%. This again shows that students have a large variation in embedding AI-generated text in their stories; in particular, the standard deviation in the percentage of AI words used is 28% per Table 1, with a coefficient of variation very close to 1, indicating a large variation.

**Table 1**

*The Utilization of AI-generated Words by Students in their Stories*

| | No. of stories with this category of chunks/words *Out of 23 stories | Average count (std. dev.) *Only stories with this category of chunks/words are included | Average length (std. dev.) *Only stories with this category of chunks/words are included |
|---|---|---|---|
| Short AI chunks | 13 | 3.77 (2.62) | 2.15 (0.72) |
| Medium AI chunks | 14 | 3.16 (2.48) | 12.38 (6.58) |
| Long AI chunks | 19 | 2.21 (1.67) | 23.27 (16.52) |
| AI chunks | 21 | 6.67 (4.99) | 13.83 (12.71) |
| Human chunks | 23 | 6.7 (4.9) | |
| AI words | 21 | 81.57 (90.56) | |
| Human words | 23 | 248.57 (164.69) | |
| Words | 23 | 323.04 (159) | |
| Percentage of AI words | | 27.95% (27.91%) | |

**Figure 5**

*The Distribution of the Percentage of AI Words among the 23 Stories*

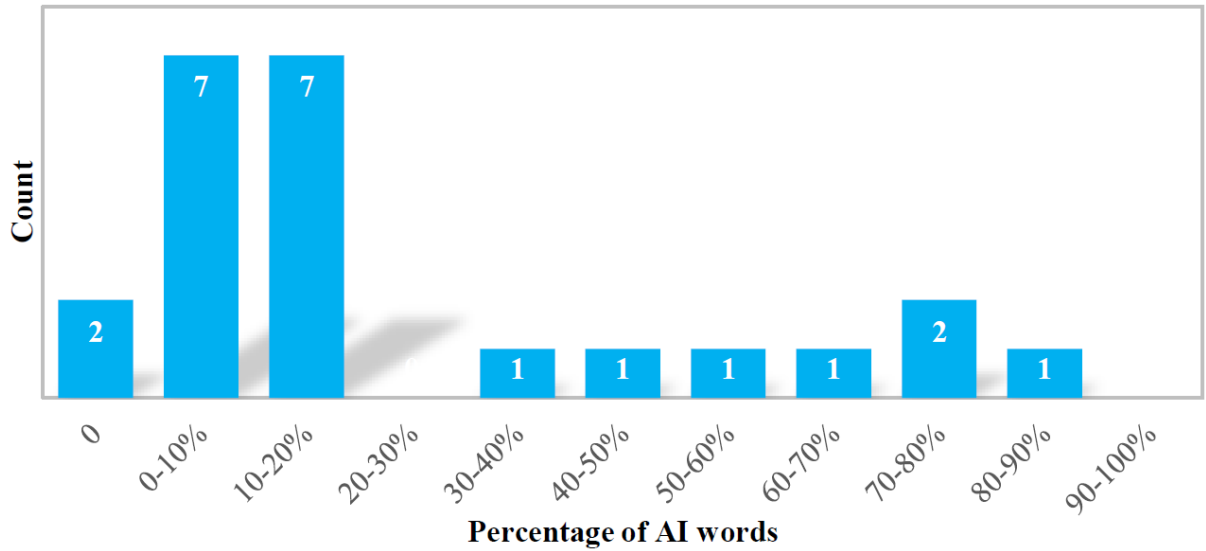

Other than students' usage of AI-generated text, we examined the scores of their stories. Table 2 shows the average scores for C, L and O, as well as the average CLO scores, AI words scores, and the grand total scores. As shown in Table 2, students scored roughly equally in the C, L, O as well as the AI words categories. Figure 6 shows the distribution of the grand total score. The scores spread widely across the whole range from 0 to the full mark of 20, with an average score of 11.2 and a standard deviation of 5.9. The corresponding coefficient of variation is 53%, and these results implied that the markers consider that there is a large variation in students' performance.

**Table 2**

*The scores in different scoring items averaged over the 23 stories*

| Scoring item | Full mark | Average mark | Standard deviation |
|---|---|---|---|
| Content (C) | 5 | 3.22 | 1.8 |
| Language (L) | 5 | 3.43 | 1.42 |
| Organization (O) | 5 | 3.07 | 1.6 |
| CLO score | 15 | 9.72 | 4.74 |
| AI words score | 5 | 3.4 | 1.84 |
| Grand total | 20 | 11.2 | 5.94 |

**Figure 6**

*The Distribution of the Grand Total Scores among the 23 Stories*

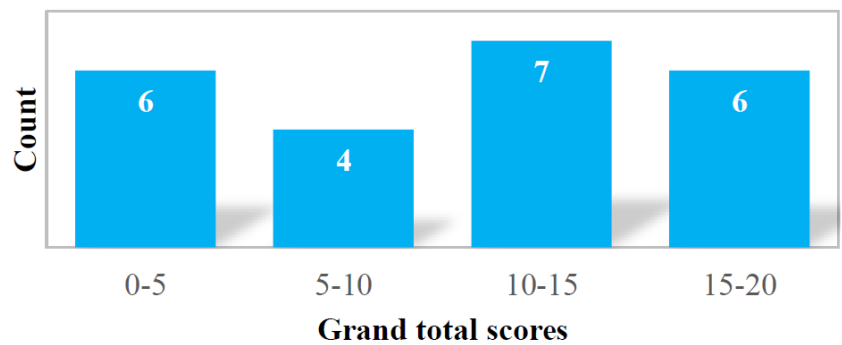

## Phase 2: Multiple Linear Regression Analysis

To identify the patterns of students' incorporation of AI-generated text in their stories and AI's contributions to the human-rated scores they obtained, we examined the following multiple linear regression (MLR) model (Aiken et al., 2003) given by Equation (1) to identify the potential correlation of various AI-generated chunks and words with different scoring items including the C, L, O, CLO, AI words and the grand total scores. Instead of multiple linear regression models, computing merely the Pearson correlation between the scores and the variables does not accurately represent the casualty relation between them, as they may be simultaneously affected by other variables and are not the cause and effect to each other as suggested by a high Pearson correlation coefficient. The multiple linear regression model can eliminate such confounding impact and reveal the dependence of scores on individual factors:

$$Score = m_1 x_1 + m_2 x_2 + m_3 x_3 + m_4 x_4 + m_5 x_5 + m_6 x_6 + C,$$ Equation (1) where $x_1$ = number of short AI chunks, $x_2$ = number of medium AI chunks, $x_3$ = number of long AI chunks, $x_4$ = number of human chunks, $x_5$ = number of AI words, and $x_6$ = number of human words.

Although a linear relationship may not fully capture the causality relationship between scores and variables, but as shown in the last column of Table 3, the model is highly statistically significant in explaining various scores, and later on give us interesting interpretations in how the scores of stories depends on the incorporation of AI-generated text. To avoid the phenomenon called co-linearity which may mask the correlation of related factors with the scores, we have not included the percentage of AI words in the model as it is related to the number of AI words and human words in the stories.

By the method of least squares (Dekking et al., 2005), one can obtain the best fitted parameters $m_i$'s in the model, but the values of $m_i$ may not fully represent the correlation of the variable $x_i$ with scores since $x_i$'s are not normalized. We thus compute the so-called partial correlations (Brown & Hendrix, 2005), which are correlations between the score and the variables after removing the co-dependence on all other variables in Equation 1. For instance, the partial correlation $\tilde{C}(score, x_1)$ between the score and $x_1$ is related to the fitted parameter $m_1$ through the following relation:

$$\tilde{C}(score, x_1) = m_1 \times \frac{residual(x_1)_{x_1=f(x_2,...,x_6)}}{residual(score)_{score=f(x_2,...,x_6)}},$$

where $residual(x_1)_{x_1=f(x_2,...,x_6)}$ and $residual(score)_{score=f(x_2,...,x_6)}$ correspond to the residuals of the linear regression models in fitting x_1 and the score respectively and separately using only the factors $x_2, ..., x_6$ (Dekking et al., 2005), and hence the co-dependence on factors $x_2, ..., x_6$ are eliminated in the partial correlation $\tilde{C}(score, x_1)$.

Table 3 shows the results of partial correlation between various scoring items and the variables $x_1, ..., x_6$. Except AI word scores, all scores are strongly positively correlated with the number of human words, with a value of $\tilde{C} > 0.9$, and all scores are also positively correlated with the number of AI words, with a value of $\tilde{C} \approx 0.5$, and these partial correlations are all statistically significant. These results imply that the more the words in the stories, regardless of whether they are written by human or AI, the higher the C, L, O scores the students obtained. As the contest limited all stories to 500 words or less, these results imply that more competent students will use more words, regardless of their own words or AI-generated text, to complete their stories and score higher. With more human and AI words, there are more occasions in the stories where AI texts have to connect to or be embedded in the human texts, or vice versa. We also note that although the scores are more strongly correlated and statistically significant with the number of human words, the number of AI words also plays its role in contributing to all C, L, O scores significantly; interestingly, the partial correlations between the number of human words and all C, L, O scores are around 0.9, while the partial correlations between the number of AI words and all C, L, O scores are around 0.5, implying both contribute roughly evenly to each of the C, L, O scores. Per existing research that a competent writer has capacity to edit and to improve so that a completed text might be very different from a first draft (Flower & Hayes, 1981), this may imply that more competent writers not only write more words, but they can edit AI words or inspiration into their own writing which contributes to C, L and O criteria. Specifically, since competent writers used more human and AI chunks in their stories and wrote more coherently, they likely repeatedly edited their compositions.

One unexpected observation is that AI word scores were significantly positively and negatively correlated with the number of human chunks and the number of medium AI chunks, respectively. This is because one may expect the opposite because AI word scores are literally assessed by AI chunks instead of human chunks. To better understand this result, we note that there are three performance levels to achieve for the AI Words criterion. For students who want to achieve the second highest performance level, they may decide to embed human chunks in longer AI chunks and avoid using medium AI chunks to increase the number of AI chunks. Hence their positive and the negative partial correlations with AI words score. To further understand this conjecture, the Pearson correlations between the number of short, medium and long AI chunks with the grand total scores of the stories are computed and are respectively 0.9, 0.03, 0.44 respectively. As we mentioned before, the values of Pearson correlations are affected by co-dependence on other variables, but here it may give us an intuitive understanding. In this case, the grand total scores of the stories seem to be uncorrelated with the number of medium AI chunks. In addition, these results of which AI word scores or positively and negatively correlated with the number of human and medium AI chunks respectively may imply that students were well aware of the competition rules on how AI word scores are assessed, and developed strategies for the efficient use of AI words in their stories.

**Table 3**

*The Partial Correlations between Various Scores and Factors*

|  | No. short AI chunks | No. med. AI chunks | No. long AI chunks | No. of human chunks | No. of AI words | No. of Human words | Model p-value |
|---|---|---|---|---|---|---|---|
| Content (C) score | 0.125 | 0.124 | 0.18 | -0.124 | 0.49* | 0.937** | $7.47 \times 10^{-8}$ |
| Language (L) score | 0.218 | 0.228 | 0.215 | -0.238 | 0.537* | 0.914** | $1.03 \times 10^{-6}$ |
| Organization (O) score | 0.075 | -0.002 | 0.172 | -0.055 | 0.532* | 0.943** | $3.99 \times 10^{-8}$ |
| CLO score | 0.168 | 0.144 | 0.224 | -0.168 | 0.586* | 0.951** | $1.16 \times 10^{-8}$ |
| AI words score | -0.13 | -0.477* | -0.357 | 0.48* | 0.089 | -0.204 | $2.15 \times 10^{-5}$ |
| Grand total | 0.056 | -0.174 | -0.032 | 0.159 | 0.5* | 0.907** | $4.29 \times 10^{-8}$ |

*Note*. A single star (*) and a double star (**) correspond to the cases of statistical significance with a p-value less than 0.05 and 0.01 respectively. The last column shows the p-value of the corresponding linear regression model in explaining the various scores by the 6 factors.

**Phase 3: Cluster Analysis**

We used cluster analysis to group students according to the language features of AI-generated text in their stories and the stories' human-rated scores. Our cluster analysis algorithms considered the following language features and scores: the number of AI chunks (short, medium, and long), the percentage of AI-generated words used in a story, the total number of words in a story, the C score, the L score, the O score, the CLO score, and finally the AI Words score. Moreover, we used different algorithms to generate different types of clusters.

*Algorithm 1: EM-Algorithm*

EM-Algorithm is an algorithm that uses Gaussian distribution to probabilistically estimate the likelihood of data points belonging to a particular cluster. The algorithm is configured to cluster the dataset into eight groups of students and the output is described in Figure 7.

**Figure 7**

*The Outputs from EM-Algorithm*

```
EExpectation-Maximization (EM) Algorithm
Cluster : 0
    No. short chunks  No. med. chunks  No. long chunks  % of AI words  Average C  Average L  Average O  AI words score  No. of words  Avg CLO  total score  Cluster ID  student_ID
0                 3                3                2          14.03        5.0        5.0        5.0               5           449     15.0         20.0           0           5
1                 8                4                1          10.60        5.0        5.0        5.0               5           500     15.0         20.0           0           7
2                 4                2                2           9.02        5.0        5.0        5.0               5           488     15.0         20.0           0           8
4                 6                7                1          19.60        5.0        5.0        4.0               5           449     14.0         19.0           0           2
6                 7                2                1          10.40        5.0        4.0        4.5               5           500     13.5         18.5           0           9

Cluster : 1
    No. short chunks  No. med. chunks  No. long chunks  % of AI words  Average C  Average L  Average O  AI words score  No. of words  Avg CLO  total score  Cluster ID  student_ID
8                 0                1                0           1.78        5.0        4.0        4.0               0           394     13.0         13.0           1           1
11                0                0                1           3.17        4.0        4.0        4.0               0           347     12.0         12.0           1          14
13                0                2                0           7.69        3.0        3.0        3.5               0           286      9.5          9.5           1          16

Cluster : 2
    No. short chunks  No. med. chunks  No. long chunks  % of AI words  Average C  Average L  Average O  AI words score  No. of words  Avg CLO  total score  Cluster ID  student_ID
15                0                1                4          77.40        1.0        2.0        1.5               0           208      4.5          4.5           2           4
17                1                1                1          61.21        1.0        2.0        1.0               1           116      4.0          5.0           2           3
18                0                0                3          56.21        1.0        2.0        1.0               0           153      4.0          4.0           2           1

Cluster : 3
    No. short chunks  No. med. chunks  No. long chunks  % of AI words  Average C  Average L  Average O  AI words score  No. of words  Avg CLO  total score  Cluster ID  student_ID
10                3                9                5          49.3         4.5        4.0        4.0               3           359     12.5         15.5           3           7

Cluster : 4
    No. short chunks  No. med. chunks  No. long chunks  % of AI words  Average C  Average L  Average O  AI words score  No. of words  Avg CLO  total score  Cluster ID  student_ID
14                1                2                0          16.20        2.0        3.0        1.5               0           179      6.5          6.5           4          11
19                0                1                0           4.38        1.0        2.0        1.0               0           137      4.0          4.0           4           6
20                0                0                0           0.00        1.0        2.0        1.0               0           137      4.0          4.0           4          10
21                1                1                1          31.16        1.0        1.0        1.5               1           138      3.5          4.5           4           2
22                0                0                0           0.00        1.0        1.0        1.0               0            29      3.0          3.0           4           5

Cluster : 5
    No. short chunks  No. med. chunks  No. long chunks  % of AI words  Average C  Average L  Average O  AI words score  No. of words  Avg CLO  total score  Cluster ID  student_ID
5                 0                1                6          88.94        4.5        5.0        4.5               0           470     14.0         14.0           5           4

Cluster : 6
    No. short chunks  No. med. chunks  No. long chunks  % of AI words  Average C  Average L  Average O  AI words score  No. of words  Avg CLO  total score  Cluster ID  student_ID
3                 0                2                2          13.54        5.0        5.0        4.5               0           495     14.5         14.5           6           3
7                 2                3                0           9.71        4.5        5.0        4.0               0           453     13.5         13.5           6          13
9                 0                6                0          17.02        4.5        4.5        4.0               0           470     13.0         13.0           6          15
12                3                5                0           8.77        4.0        3.5        4.0               0           479     11.5         11.5           6          12

Cluster : 7
    No. short chunks  No. med. chunks  No. long chunks  % of AI words  Average C  Average L  Average O  AI words score  No. of words  Avg CLO  total score  Cluster ID  student_ID
16                1                7                2          76.8         1.0        2.0        1.0               3           194      4.0          7.0           7           6
```

First, we observe students in clusters *Zero* and *Six* received a full or near full CLO score. Additionally, AI-generated words made up only 9% to 19% of their total word count, which is low compared to other clusters. The stories written by students in these two clusters also tend to be longer, nearer the 500-word limit, compared to other clusters. This indicates that the most competent writers demonstrate less inclination to seek assistance from AI, which confirms results from descriptive statistics and MLR.

Although students in these two clusters appeared to be competent writers that relied less on AI-generated text, we observed a difference in these clusters' AI Words scores. On the one hand, students in cluster Zero performed quite well in the AI Words criterion. On the other hand, all students in clusters Six received no score for the AI Words criterion. This may indicate these students' ignorance or misunderstanding about the scoring rubric descriptors; or these students intentionally did not use AI-generated text according to the scoring rubric descriptors.

Among clusters of high users of AI-generated text, we observed two different results. Students in clusters Three and Five wrote stories with AI-generated words comprising 50% to 89% of the total number of words. Compared to other students, they scored relatively high on their CLO (80th percentile), specifically, 4.5 for C, 4.5 for O and 4.25 for L. On the other hand, cluster Three produced stories with AI-generated words comprising 55% to 78% of the total number of words but the students' CLO score was low; specifically, they scored 1 point for C and O criteria, and 2 points for the L criteria. The CLO scores of clusters Two and Seven indicate that those students did not submit complete stories, for which reason they could not exceed 1 point for the C criteria but could achieve an additional point for L.  We interpret clusters One and Seven as students who did not appear to be competent writers, had decided to seek more support from their AI-NLG tools but had not developed effective strategies to leverage AI-generated text in their writing. In this way, our study provides evidence to question claims that compositions comprising exclusively AI-generated text would be scored positively for cohesiveness and grammatical accuracy (Godwin-Jones, 2022), for which reason automatic writing evaluation systems would have limited value. On the other hand, the results suggest that students in clusters Three and Five were more competent writers, had decided to seek more assistance from their tools, and had developed effective strategies to incorporate more AI-generated text in their stories.

## *Algorithm 2: K-Means Clustering*

K-Means Clustering statistically identifies the K number of clusters in the dataset by measuring the distance between each data to the center of each cluster (see Figure 8). Like with the EM-Algorithm clustering analysis, we configured K to 8 clusters. K-means Clustering revealed additional insights into students' interaction patterns with AI-generated text – the attribution of specific language features of AI-generated text in stories to human-rated scores.

**Figure 8**

*The Outputs from K-Means Algorithm*

```
K-Means Clustering Algorithm
Cluster : 0
    No. short chunks  No. med. chunks  No. long chunks  % of AI words  Average C  Average L  Average O  AI words score  No. of words  Avg CLO  total score  Cluster ID  student_ID
13               0                2                0           7.69        3.0        3.0        3.5               0           286      9.5          9.5           0          16

Cluster : 1
    No. short chunks  No. med. chunks  No. long chunks  % of AI words  Average C  Average L  Average O  AI words score  No. of words  Avg CLO  total score  Cluster ID  student_ID
0                3                3                2          14.03        5.0        5.0        5.0               5           449     15.0         20.0           1           5
1                8                4                1          10.60        5.0        5.0        5.0               5           500     15.0         20.0           1           7
2                4                2                2           9.02        5.0        5.0        5.0               5           488     15.0         20.0           1           8
3                0                2                2          13.54        5.0        5.0        4.5               0           495     14.5         14.5           1           3
4                6                7                1          19.60        5.0        5.0        4.0               5           449     14.0         19.0           1           2
6                7                2                1          10.40        5.0        4.0        4.5               5           500     13.5         18.5           1           9
7                2                3                0           9.71        4.5        5.0        4.0               0           453     13.5         13.5           1          13
9                0                6                0          17.02        4.5        4.5        4.0               0           470     13.0         13.0           1          15
12               3                5                0           8.77        4.0        3.5        4.0               0           479     11.5         11.5           1          12

Cluster : 2
    No. short chunks  No. med. chunks  No. long chunks  % of AI words  Average C  Average L  Average O  AI words score  No. of words  Avg CLO  total score  Cluster ID  student_ID
17               1                1                1          61.21        1.0        2.0        1.0               1           116      4.0          5.0           2           3
18               0                0                3          56.21        1.0        2.0        1.0               0           153      4.0          4.0           2           1
21               1                1                1          31.16        1.0        1.0        1.5               1           138      3.5          4.5           2           2

Cluster : 3
    No. short chunks  No. med. chunks  No. long chunks  % of AI words  Average C  Average L  Average O  AI words score  No. of words  Avg CLO  total score  Cluster ID  student_ID
15               0                1                4          77.4         1.0        2.0        1.5               0           208      4.5          4.5           3           4
16               1                7                2          76.8         1.0        2.0        1.0               3           194      4.0          7.0           3           6

Cluster : 4
    No. short chunks  No. med. chunks  No. long chunks  % of AI words  Average C  Average L  Average O  AI words score  No. of words  Avg CLO  total score  Cluster ID  student_ID
22               0                0                0           0.0         1.0        1.0        1.0               0            29      3.0          3.0           4           5

Cluster : 5
    No. short chunks  No. med. chunks  No. long chunks  % of AI words  Average C  Average L  Average O  AI words score  No. of words  Avg CLO  total score  Cluster ID  student_ID
8                0                1                0           1.78        5.0        4.0        4.0               0           394     13.0         13.0           5           1
10               3                9                5          49.30        4.5        4.0        4.0               3           359     12.5         15.5           5           7
11               0                0                1           3.17        4.0        4.0        4.0               0           347     12.0         12.0           5          14

Cluster : 6
    No. short chunks  No. med. chunks  No. long chunks  % of AI words  Average C  Average L  Average O  AI words score  No. of words  Avg CLO  total score  Cluster ID  student_ID
14               1                2                0          16.20        2.0        3.0        1.5               0           179      6.5          6.5           6          11
19               0                1                0           4.38        1.0        2.0        1.0               0           137      4.0          4.0           6           6
20               0                0                0           0.00        1.0        2.0        1.0               0           137      4.0          4.0           6          10

Cluster : 7
    No. short chunks  No. med. chunks  No. long chunks  % of AI words  Average C  Average L  Average O  AI words score  No. of words  Avg CLO  total score  Cluster ID  student_ID
5                0                1                6          88.94        4.5        5.0        4.5               0           470     14.0         14.0           7           4
```

Students in cluster One wrote the longest stories of at least 449 words and AI-generated words comprised only 19% of their stories' words. However, less than half of the students in this cluster achieved a full CLO score of 15. This indicates that the use of AI-generated text does not necessarily contribute to the highest CLO performance levels.

Another observation is that students in cluster Six scored low on CLO and on AI Words and wrote the least number of words in their stories, 157 words on average, compared to all students. The low CLO scores indicate that students did not submit complete stories. Additionally, they had not used AI-generated text to extend their writing. The implication is that less competent writers may lack the editing ability and strategies by which they might use AI-generated text to complete writing tasks.

*Algorithm 3: Mean-Shift Clustering*

Different from EM-Algorithm and K-Means Clustering, Mean-Shift Clustering does not require a predetermined number of clusters. Instead, it determines the number of clusters in a dataset and assigns each data point to the clusters by shifting points towards the highest density (or mean value) in each cluster (see Figure 9).  This algorithm grouped the students into six clusters.

**Figure 9**

*The Outputs from Means-Shift Algorithm*

```
Mean Shift Algorithm
Cluster : 0
    No. short chunks  No. med. chunks  No. long chunks  % of AI words  Average C  Average L  Average O  AI words score  No. of words  Mean Shift  AI Score  total score  Cluster ID  student_ID
0                  3                3                2          14.03        5.0        5.0        5.0               5           449           0         5         20.0           0           5
1                  8                4                1          10.60        5.0        5.0        5.0               5           500           0         5         20.0           0           7
2                  4                2                2           9.02        5.0        5.0        5.0               5           488           0         5         20.0           0           8
3                  0                2                2          13.54        5.0        5.0        4.5               0           495           0         0         14.5           0           3
4                  6                7                1          19.60        5.0        5.0        4.0               5           449           0         5         19.0           0           2
6                  7                2                1          10.40        5.0        4.0        4.5               5           500           0         5         18.5           0           9
7                  2                3                0           9.71        4.5        5.0        4.0               0           453           0         0         13.5           0          13
9                  0                6                0          17.02        4.5        4.5        4.0               0           470           0         0         13.0           0          15
12                 3                5                0           8.77        4.0        3.5        4.0               0           479           0         0         11.5           0          12

Cluster : 1
    No. short chunks  No. med. chunks  No. long chunks  % of AI words  Average C  Average L  Average O  AI words score  No. of words  Mean Shift  AI Score  total score  Cluster ID  student_ID
14                 1                2                0          16.20        2.0        3.0        1.5               0           179           1         0          6.5           1          11
15                 0                1                4          77.40        1.0        2.0        1.5               0           208           1         0          4.5           1           4
16                 1                7                2          76.80        1.0        2.0        1.0               3           194           1         3          7.0           1           6
17                 1                1                1          61.21        1.0        2.0        1.0               1           116           1         1          5.0           1           3
18                 0                0                3          56.21        1.0        2.0        1.0               0           153           1         0          4.0           1           1
19                 0                1                0           4.38        1.0        2.0        1.0               0           137           1         0          4.0           1           6
20                 0                0                0           0.00        1.0        2.0        1.0               0           137           1         0          4.0           1          10
21                 1                1                1          31.16        1.0        1.0        1.5               1           138           1         1          4.5           1           2

Cluster : 2
    No. short chunks  No. med. chunks  No. long chunks  % of AI words  Average C  Average L  Average O  AI words score  No. of words  Mean Shift  AI Score  total score  Cluster ID  student_ID
8                  0                1                0           1.78        5.0        4.0        4.0               0           394           2         0         13.0           2           1
10                 3                9                5          49.30        4.5        4.0        4.0               3           359           2         3         15.5           2           7
11                 0                0                1           3.17        4.0        4.0        4.0               0           347           2         0         12.0           2          14

Cluster : 3
    No. short chunks  No. med. chunks  No. long chunks  % of AI words  Average C  Average L  Average O  AI words score  No. of words  Mean Shift  AI Score  total score  Cluster ID  student_ID
13                 0                2                0           7.69        3.0        3.0        3.5               0           286           3         0          9.5           3          16

Cluster : 4
    No. short chunks  No. med. chunks  No. long chunks  % of AI words  Average C  Average L  Average O  AI words score  No. of words  Mean Shift  AI Score  total score  Cluster ID  student_ID
5                  0                1                6          88.94        4.5        5.0        4.5               0           470           4         0         14.0           4           4

Cluster : 5
    No. short chunks  No. med. chunks  No. long chunks  % of AI words  Average C  Average L  Average O  AI words score  No. of words  Mean Shift  AI Score  total score  Cluster ID  student_ID
22                 0                0                0           0.00        1.0        1.0        1.0               0            29           5         0          3.0           5           5
```

One observation was that the use rate of AI-words varied greatly, from 0 to 77.4%, in cluster One, which featured students with the lowest CLO scores and the shortest stories. This indicates that the length of a story is a more accurate indicator of a student's writing competence than the number of AI-words used in a story. Additionally, a students' higher use of AI-generated text might improve the student's CLO score if the student were to write nothing at all without AI-generated text. Therefore, this cluster provides an initial clue that the use of AI-generated text does not compensate for all students' writing competence.

To summarize, we plot three-dimensional graphs of the results for each cluster analysis algorithm, showing a horizontal view in Figure 10 and a vertical view in Figure 11. From these three-dimensional graphs, visually the students can be categorized into four learner profiles, 1) students with high CLO scores, but fewer AI words; 2) students with high scores, but more AI words; 3) students with low scores, but fewer AI words; and 4) students with low scores, but more AI words.

**Figure 10**

*Horizontal View of Three Clustering Algorithm Three-dimensional Graphs*

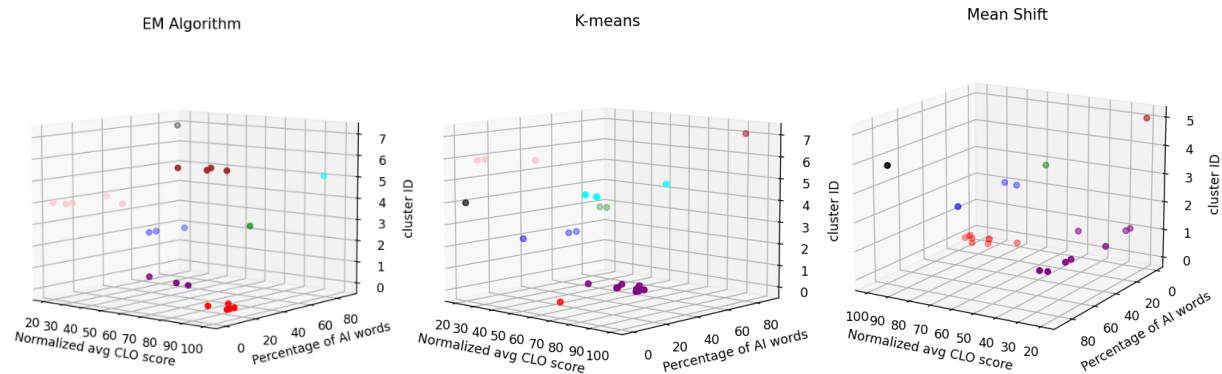

*Note*. The x-axis is the normalized average CLO score between 0 and 100; the y-axis is the percentage of AI words used in the writing; and the z-axis is the cluster ID.

**Figure 11**

*Vertical View of Three Clustering Algorithm Three-dimensional Graphs*

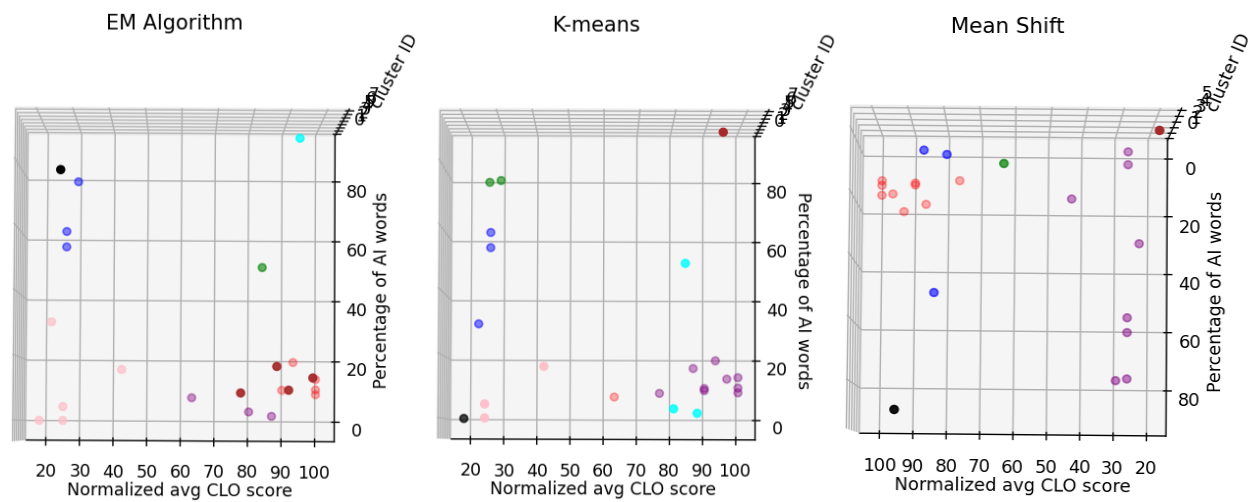

## Conclusion

**Major Findings**

Our writing with a machine-in-the-loop study explored EFL students' usage of AI-generated text in written compositions, focusing on how that text may affect human-rated scores of compositions. The descriptive statistics show a large variation in students' use of AI-generated text in their stories. At the same time, human scoring indicates a large variation in the C, L and O scores of students' stories.

We aimed to identify any patterns of how students have strategically edited AI-generated text into their compositions and whether any patterns affect scores on the overall quality of the work. A multiple linear regression of AI-generated text language features and CLO scores show the number of human words and the number of AI-generated words contribute significantly to CLO scores. Significant correlations between different syntactic forms of AI-generated text and AI Words scores highlight a group of competent and strategic students who use AI-generated text as a means to boost their AI Words scores.

Clustering analyses of students reveal distinct learner profiles: those competent writers who strategically employ specific syntactic forms of AI-generated text but use less AI-generated text compared to their peers, and those that effectively utilize more AI-generated text. Notably, our analyses consistently indicate that EFL students with higher proficiency writing skills demonstrate a reduced reliance on AI-generated text while being better able to edit that text to achieve higher performance levels. Their editing resembles that of adult writers who prefer editing AI-generated text when writing with a machine-in-the-loop (Calderwood et al., 2020; Yang et al., 2022). The association between editing and writing performance also resembles

Kangasharju et al.'s (2022) suggestion of an association between the quality of a student's poem and the number of edits made on a draft of that poem from an AI-NLG tool.

Equally important, our investigation highlights that even when students have access to AI-NLG tools and their generative text, they may not utilize these to their maximum potential. This finding resembles those from adult populations that have produced mixed outcomes when writing with a machine-in-the-loop (Clark et al., 2018). Our finding suggests that the level of AI-word usage varies depending on a student's existing writing ability, with more proficient students relying less on AI-words. Furthermore, a high level of AI-word usage does not guarantee high performance in writing outcomes: we observed instances where students who used AI-words extensively still performed poorly, such as writing significantly shorter essays (around 150 words out of a 500-word limit) and receiving low human-rated scores. Even when writing with a machine-in-the-loop, prolifically using AI-words, they demonstrate writing ability typical of EFL students (Hyland, 2003a).

Nonetheless, a closer examination of individual scores for C, L, and O reveals that AI-generated text may improve the quality of writing for low-scoring EFL students, particularly in the language category. We attribute this advantage in the language category to the advancements in AI-NLG tools, which are capable of generating proper sentences with correct grammar (Dizon & Gayed, 2021).

The analyses also reveal learner profiles of less competent writers, those that use little AI-generated text and may not have strategies for its effective use (Crossely et al., 2016); and those that use more AI-generated text and that benefit from its use if they were to otherwise write nothing at all.  The latter EFL student profile resembles Gayed et al.'s (2022) sample of adult EFL students who could output a greater number of words with AI-generated text, although that greater number did not necessarily improve the quality of writing. For these reasons, our analysis provides an initial clue that the usage of AI-generated text does not mask students' existing writing competency.

In sum, our study has evidenced that students' writing with a machine-in-the-loop does not preclude students using AI-generated text much in their writing and students using AI-generated text to realize higher levels of writing performance. Importantly, although AI-NLG tools may be one way to address EFL students' challenges in the writing classroom, our study has evidenced that students' writing with a machine-in-the-loop does not decisively benefit all students' writing.

**Pedagogical Implications**

Our study can inform the flexible design of AI-related curricula to address the various needs of schools and students, especially for secondary education for which there has been little research (Chiu, 2021). First, our study contributes a curricular effort for EFL students to write with a machine-in-the-loop. Specifically, as writing activities with AI-NLG tools have been scarce (Lin & Chang, 2020), our study contributes design specifications for an authentic writing task and AI-NLG tools that can be freely replicated and adapted. Second, we provide replicable metrics for assessing the basic structure, organization, and syntactic complexity of AI-generated text. These measures can inform rubrics and complement other assessment modes of student compositions written with AI-generated text.

Furthermore, our findings can inform teachers' instructional methods for different groups of EFL students to complete writing tasks with a machine-in-the-loop. More competent writers are more able to use AI-NLG tools in a supporting role and to edit AI-generated text to complete a high-quality composition. These writers can benefit from fine-grain instruction on different phases of writing with a machine-in-the-loop, for instance, on ways to prompt AI-NLG tools so as to unlock different roles for AI-NLG tools in the writing process and methods to evaluate AI-generated text that may enhance writing performance. For less competent students, they will need additional support from teachers, peers, and other tools in their writing classroom to enhance their writing competence. To achieve higher levels of writing performance, they may require more fundamental instruction on genre and methods to generate words to complete a written composition, besides instruction on granular editing of AI-generated text and modeling of strategic and effective integration of AI-generated text in a composition. Because of differentiated  instructional methods, students at different competence levels might write with a machine-in-the-loop so as to enhance their writing.

Importantly, some universities and institutions have been banning their students' use of AI-generated text for classroom, coursework, and assessment tasks, which might be in part due to an unclear picture of how AI-generated text can be used in an effective and ethically correct way. Our empirical evidence shows that it is essential to continue scrutinizing the pedagogical value of AI-generated text in education.

**Limitations and Future Research**

For our writing with a machine-in-the-loop study, we operationalized relatively simple measures of basic structure and organization and syntactic complexity to analyze AI-generated text in compositions. Future studies may explore quality writing with AI-generated text by operationalizing measures for other language features that predict human judgements of both first language and EFL writing proficiency: lexical sophistication (i.e., the use of advanced words that indicate lexical knowledge, often measured by number-of-word calculations); and text cohesion (i.e., the interconnectivity of text segments) ([Crossley, 2020](#)); future studies can analyze AI-generated text in terms of discourse features. With further study with additional measures, we can better understand how competent writers have effectively integrated larger quantities of AI-generated text into their writing.

In our study, students composed short stories and in subsequent research can expand students' writing to argumentative and factual text types. Furthermore, our study's AI-NLG tools utilized open-source, free-to-use language models and could output at most a paragraph of text. For the tools to better meet learner's needs (Vygotsky, 1978), educators could fine-tune such models on different corpora, including bodies of student compositions, to improve their capacity to generate relevant language for the target text type and explore such models' contributions to student writing. Besides, study should expand to the largest, proprietary language models like ChatGPT and to AI-NLG tools that can generate several paragraphs of text.

Although our study evidenced that AI-NLG tools that predict text show some potential to improve students' writing, the tools suffer limitations and using them raises risk in the writing classroom. One limitation is such tools' propensity to *hallucinate*, that is, to generate text that deviates from its source input and fails to meet user expectations (Ji et al., 2022). Such text can comprise offensive or biased material (Bender et al., 2021), or factually incorrect and nonsensical answers, although plausible sounding (OpenAI, 2022). Another limitation is such tools' propensity to *degenerate*, that is, to generate bland and repetitive text (Holtzman et al., 2020). Since rewriting a prompt can significantly change a tool's performance, further research into writing with a machine-in-the-loop will be necessary into prompting (Reynolds & McDonell, 2021) AI-NLG tools in the writing classroom so that a studently careful selects input and controls an AI-NLG tool to perform a desired task. Besides, future studies could investigate the writing approaches by which a student would prompt an AI-NLG tool, what prompts students have used to generate text, and whether students perceive any AI-generated text as satisfactory for integration into a composition. Different methods may capture students' interactions with AI-NLG tools in the writing process, such as screen recordings, think-aloud protocols, interviews, and stimulated recalls, to provide a complete understanding of how the tools contribute to students' writing and how different students use AI for writing.

EXPLORING AI-GENERATED TEXT IN STUDENT WRITING                                          38

EXPLORING AI-GENERATED TEXT IN STUDENT WRITING 40

EXPLORING AI-GENERATED TEXT IN STUDENT WRITING                                                41

# Appendix A

*The Assessment Rubric for the 1st Human-AI Creative Writing Contest for Hong Kong Secondary Schools*

| Score | Content | Language | Organization | AI Words |
|---|---|---|---|---|
| 5 | · Content fulfills the requirements of the question<br>· Almost totally relevant<br>· Most ideas are well developed/supported<br>· Creativity and imagination are shown when appropriate<br>· Shows general awareness of audience | · Wide range of accurate sentence structures with a good grasp of simple and complex sentences<br>· Grammar mainly accurate with occasional common errors that do not affect overall clarity<br>· Vocabulary is wide, with many examples of more sophisticated lexis<br>· Spelling and punctuation are mostly correct<br>· Register, tone and style are appropriate to the genre and text-type | · Text is organized effectively, with logical development of ideas<br>· Cohesion in most parts of the text is clear<br>· Strong cohesive ties throughout the text<br>· Overall structure is coherent, sophisticated and appropriate to the genre and text-type | · AI words compose less than ⅓ of the total number of words in the text |
| 3 | · Content just satisfies the requirements of the question<br>· Relevant ideas but may show some gaps or redundant information<br>· Some ideas but not well developed<br>· Some evidence of creativity and imagination<br>· Shows occasional awareness of audience | · Simple sentences are generally accurately constructed.<br>· Occasional attempts are made to use more complex sentences. Structures used tend to be repetitive in nature<br>· Grammatical errors sometimes affect meaning<br>· Common vocabulary is generally appropriate<br>· Most common words are spelt correctly, with basic punctuation being accurate<br>· There is some evidence of register, tone and style appropriate to the genre and text-type | · Parts of the text have clearly defined topics<br>· Cohesion in some parts of the text is clear<br>· Some cohesive ties in some parts of the text<br>· Overall structure is mostly coherent and appropriate to the genre and text-type | · At least 8 AI chunks of any length |
| 1 | · Content shows very limited attempts to fulfill the requirements of the question<br>· Intermittently relevant; ideas may be repetitive<br>· Some ideas but few are developed<br>· Ideas may include misconception of the task or some inaccurate information<br>· Very limited awareness of audience | · Some short simple sentences accurately structured<br>· Grammatical errors frequently obscure meaning<br>· Very simple vocabulary of limited range often based on the prompt(s)<br>· A few words are spelt correctly with basic punctuation being occasionally accurate | · Parts of the text reflect some attempts to organize topics<br>· Some use of cohesive devices to link ideas | · AI words used in long chunks (more than 1 sentence in length) and in short chunks (less than 5 words in length). |

*Note*.
1. Content mark cannot exceed 1 if the story is not complete, that is, missing exposition; conflict; climax; and / or resolution.
2. Content mark cannot exceed 1 if the story is not a story, for example, an article or an essay
3. Creativity in content refers to the details, transformation and originality of ideas
4. Language and organization marks cannot exceed +/- 1 of the content mark.

**Appendix B**

*Human-rated scores for CW students' stories*

| Text Name | Marker 1 | | | | Marker 2 | | | | Marker Average | | | | AI words | Grand total |
|---|---|---|---|---|---|---|---|---|---|---|---|---|---|---|
| | Content (C) | Language (L) | Organization (O) | Sub-total | C | L | O | Sub-total | C | L | O | CLO | | |
| 1 | 1 | 2 | 1 | 4 | 1 | 2 | 1 | 4 | 1 | 2 | 1 | 4 | 0 | 4 |
| 2 | 1 | 1 | 1 | 3 | 1 | 1 | 2 | 4 | 1 | 1 | 1.5 | 3.5 | 1 | 5 |
| 3 | 1 | 2 | 1 | 4 | 1 | 2 | 1 | 4 | 1 | 2 | 1 | 4 | 1 | 5 |
| 4 | 1 | 2 | 1 | 4 | 1 | 2 | 2 | 5 | 1 | 2 | 1.5 | 4.5 | 1 | 6 |
| 5 | 1 | 1 | 1 | 3 | 1 | 1 | 1 | 3 | 1 | 1 | 1 | 3 | 0 | 3 |
| 6 | 1 | 2 | 1 | 4 | 1 | 2 | 1 | 4 | 1 | 2 | 1 | 4 | 3 | 7 |
| 7 | 5 | 4 | 4 | 13 | 4 | 4 | 4 | 12 | 4.5 | 4 | 4 | 12.5 | 3 | 16 |

*Human-rated scores for HMT students' stories*

| Text Name | Marker 1 | | | | Marker 2 | | | | Marker Average | | | | AI words | Grand total |
|---|---|---|---|---|---|---|---|---|---|---|---|---|---|---|
| | Content (C) | Language (L) | Organization (O) | Sub-total | C | L | O | Sub-total | C | L | O | CLO | | |
| 1  | 5 | 4 | 4 | 13 | 5 | 4 | 4 | 13 | 5   | 4   | 4   | 13   | 0 | 13 |
| 2  | 5 | 5 | 4 | 14 | 5 | 5 | 4 | 14 | 5   | 5   | 4   | 14   | 5 | 19 |
| 3  | 5 | 5 | 5 | 15 | 5 | 5 | 4 | 14 | 5   | 5   | 4.5 | 14.5 | 0 | 15 |
| 4  | 5 | 5 | 4 | 14 | 4 | 5 | 5 | 14 | 4.5 | 5   | 4.5 | 14   | 0 | 14 |
| 5  | 5 | 5 | 5 | 15 | 5 | 5 | 5 | 15 | 5   | 5   | 5   | 15   | 5 | 20 |
| 6  | 1 | 2 | 1 | 4  | 1 | 2 | 1 | 4  | 1   | 2   | 1   | 4    | 0 | 4  |
| 7  | 5 | 5 | 5 | 15 | 5 | 5 | 5 | 15 | 5   | 5   | 5   | 15   | 5 | 20 |
| 8  | 5 | 5 | 5 | 15 | 5 | 5 | 5 | 15 | 5   | 5   | 5   | 15   | 5 | 20 |
| 9  | 5 | 4 | 4 | 13 | 5 | 4 | 5 | 14 | 5   | 4   | 4.5 | 13.5 | 5 | 19 |
| 10 | 1 | 2 | 1 | 4  | 1 | 2 | 1 | 4  | 1   | 2   | 1   | 4    | 0 | 4  |
| 11 | 3 | 4 | 2 | 9  | 1 | 2 | 1 | 4  | 2   | 3   | 1.5 | 6.5  | 0 | 7  |
| 12 | 4 | 3 | 4 | 11 | 4 | 4 | 4 | 12 | 4   | 3.5 | 4   | 11.5 | 0 | 12 |
| 13 | 5 | 5 | 4 | 14 | 4 | 5 | 4 | 13 | 4.5 | 5   | 4   | 13.5 | 0 | 14 |
| 14 | 5 | 4 | 4 | 13 | 3 | 4 | 4 | 11 | 4   | 4   | 4   | 12   | 0 | 12 |
| 15 | 5 | 5 | 4 | 14 | 4 | 4 | 4 | 12 | 4.5 | 4.5 | 4   | 13   | 0 | 13 |
| 16 | 3 | 3 | 4 | 10 | 3 | 3 | 3 | 9  | 3   | 3   | 3.5 | 9.5  | 0 | 10 |